\documentclass[fleqn,10pt]{wlscirep}
\usepackage[utf8]{inputenc}
\usepackage[T1]{fontenc}
\usepackage{url}
\usepackage{booktabs}
\usepackage{multirow}
\usepackage{adjustbox}
\usepackage{comment}

\title{Submanifold Sparse Convolutional Networks for Automated 3D Segmentation of Kidneys and Kidney Tumours in Computed Tomography}

\author[1,*]{Sa\'ul Alonso-Monsalve}
\author[2]{Leigh H. Whitehead}
\author[3]{Adam Aurisano}
\author[4]{Lorena Escudero Sanchez}
\affil[1]{ETH Zürich, Institute for Particle Physics and Astrophysics, Zürich, 8093,
Switzerland}
\affil[2]{University of Cambridge, Department of Physics, Cambridge, CB3 0HE, UK}
\affil[3]{University of Cincinnati, Department of Physics, Cincinnati, 45221-0011, OH, USA}
\affil[4]{University of Cambridge, Department of Radiology, Cambridge, CB2 0QQ, UK}
\affil[*]{salonso@ethz.ch}


\begin{abstract}
Accurate delineation of kidney tumours in Computed Tomography (CT) is essential for downstream quantitative analysis and precision oncology that could enable personalised treatments, but manual segmentation is a specialised task, time-consuming and difficult to scale in routine practice. Automated 3D segmentation remains challenging in medical imaging, where images are large and dense volumes of data, making high-resolution processing with conventional dense convolutional neural networks computationally expensive, and often reliant on downsampling or patch-based inference. To overcome this problem, we propose a two-stage 3D segmentation methodology based on voxel sparsification and submanifold sparse convolutional networks (SSCNs). In Stage~1, a low-resolution sparse network identifies a region of interest (ROI); in Stage~2, a high-resolution sparse network performs refined segmentation within the cropped ROI. This design enables native 3D processing at high resolution while reducing CPU/GPU memory usage and inference time.
We evaluate the method on the KiTS23 dataset of renal cancer CT scans using 5-fold cross-validation. Our method achieved Dice similarity coefficients of 95.8\% for kidneys + masses, 85.7\% for tumours + cysts, and 80.3\% for tumours alone, with performance competitive with top KiTS23 approaches. In direct comparisons on the same cross-validation folds, the proposed sparse method achieves tumour + cyst and tumour-only Dice scores comparable to, and slightly higher than, a patch-based nnU-Net baseline, while consistently requiring less VRAM and shorter inference time across the tested hardware. Across the tested GPUs, our sparse model is markedly faster than both nnU-Net and the zero-shot zoom-out/zoom-in foundation model SegVol, which localises kidneys well but underperforms on small heterogeneous lesions. Compared to an equivalent dense implementation of the same architecture, the proposed sparse approach achieves up to a 60\% reduction in inference time and up to a 75\% reduction in VRAM usage across both CPU and the GPU configurations tested.
\end{abstract}

\begin{document}

\flushbottom
\maketitle

\thispagestyle{empty}

\section*{Introduction}

Medical imaging is routinely used for cancer diagnosis and treatment monitoring, with Computed Tomography (CT) being one of the most common image modalities used worldwide\cite{CToncology2}. This is indeed the case for renal cancer, one of the deadliest cancers \cite{KidneyCancer}: when a kidney tumour is suspected, the patient typically undergoes a contrast-enhanced computed tomography (CECT) scan for confirmation \cite{KidneyGuidelines, KidneyManagement}.

Beside diagnosis, the quantitative analysis of CT images and the extraction of downstream measurements such as radiomics \cite{radiomics} for machine-learning algorithms is proving promising in classifying tumour types, predicting their evolution, and the response of patients to treatment \cite{KidneyRadiomics, Rundo2022}. However, the accurate delineation of lesions necessary to perform such analyses is a very time-consuming task that relies on expert radiologists, and currently presents a bottleneck to deploy advanced techniques that would enable personalised treatments~\cite{jimaging8030055} in the clinical setting. Hence the growing interest in the cancer research community to develop methods for automatic tumour segmentation \cite{KiTS19Challenge,Thomas2022,TotalSegmentator, RANJBARZADEH2023106405}. 

In particular, deep neural networks based on the U-Net architecture \cite{unet} have shown the most promising results in segmenting tumours and organs in medical images, including CT, achieving some of the state-of-the-art performance in biomedical segmentation challenges \cite{isensee2021nnunet}. 
In practice, however, representing scans as 3D images is computationally infeasible without the application of downsampling. For example, a single CT scan volume with 2D images consisting of $512\times512$ pixels and 100 2D slices would have $\sim$\,26 million voxels, tens of times larger than the number of voxels that can be stored in memory on a typical GPU card. 

In addition, a dense representation, as used in traditional convolutional neural networks (CNN) for medical imaging~\cite{kshatri2023convolutional}, is inefficient when many voxels convey no useful information for the classification task. In such cases, a sparse representation that stores the position and value of only meaningful voxels is more suitable. Submanifold sparse convolutional neural networks (SSCNs)~\cite{Graham-2017-submanifold, Graham-2017-3d} were developed specifically to operate on sparse tensors, solving this problem. In scientific fields such as high-energy physics, where sparse images are common, SSCNs have been shown to outperform standard CNNs~\cite{Domine-2020-scalable,Adams:2019uqx,NEXT:2020jmz}. But despite their promising results and advantageous reduction of computing resources, both of great importance in clinical research, SSCNs have not yet been studied in the context of cancer research and only one example of the usage of these networks exists in medical imaging at all, which was recently proposed specifically for the reconstruction of 3D skulls~\cite{Li2022, li2023sparse}. For these reasons, in this paper we studied for the first time the application of SSCNs to the semantic segmentation of tumours in CT images. 

Although CT images are not naturally sparse, this imaging modality is a suitable use case for SSCNs. This is possible since the intensities stored in voxels in CT are described by the scale of Hounsfield Units (H.U.) that represents radiodensity~\cite{HounsfieldNobel}, such that different values are indicative of different tissue types. Therefore, by selecting thresholds in the voxel intensities, we can easily discard structures not useful for the tumour segmentation task, such as bones or air.

In this paper, we propose a new methodology that deploys the use of voxel sparsification and SSCNs as an alternative for CT segmentations of organs and tumours, and we demonstrate a successful application of this methodology applied to segmenting kidneys and renal tumours in CT images. 
As demonstrated by our results, not only is this alternative approach able to obtain high accuracy, competitive with other state-of-the-art methods, but it also presents an advantage in terms of resources by requiring less computation time and memory, as will be explained in the rest of this paper.

To compare this approach to commonly used practical alternatives, we also include direct comparisons, in terms of segmentation accuracy and computational cost, with a patch-based dense segmentation framework (nnU-Net~\cite{isensee2021nnunet}) and a zoom-out/zoom-in foundation model (SegVol~\cite{du2024segvol}), using the same data splits.
These results will help in the development and adoption of automatic segmentation methods in the clinical setting in a robust and resource-conscious way, which can be extended to other organs and tumour types in the future.


\section*{Methods}

\begin{figure*}[tbh]
   \includegraphics[width=1.0\textwidth]{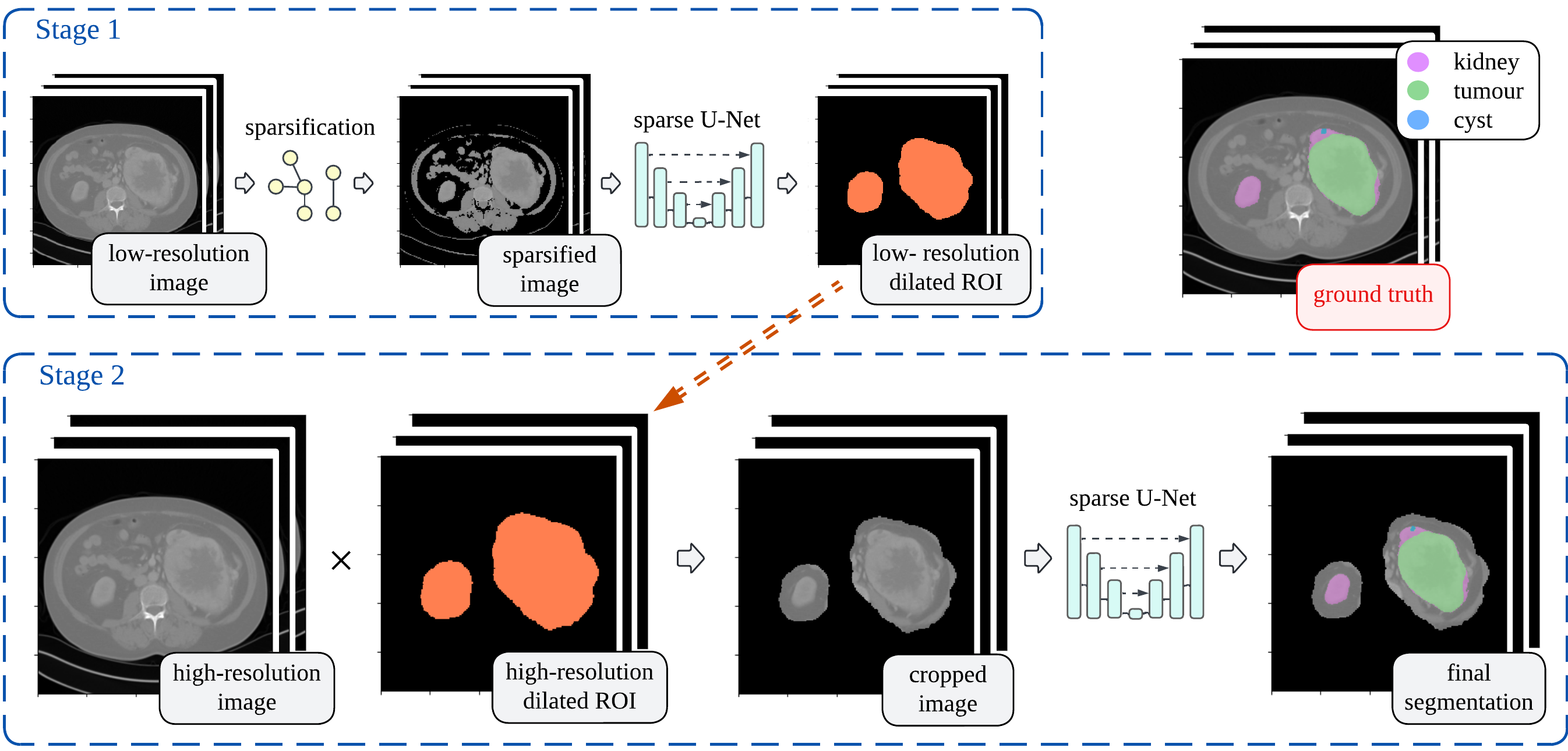}
   \caption{\textbf{Overview of the two-stage segmentation framework utilising Sparse Submanifold Convolutional Networks (SSCNs)}. In Stage 1 (ROI finder), a low-resolution sparsified image is processed by a sparse 3D U-Net to identify a region of interest (ROI). Such ROIs are then dilated to be conservative and ensure all relevant structures are included, and passed to the next stage and used to crop the original high-resolution image. In Stage 2 (Segmentation), the high-resolution cropped image is fed into another sparse 3D U-Net to obtain the final segmentations. This two-step approach efficiently reduces computational cost while maintaining segmentation accuracy by focusing on relevant anatomical structures. The segmentation outputs include kidneys, tumours and cysts, independently segmented following the ground truth annotations from the KiTS23 dataset.}
   \label{fig:method}
\end{figure*}

The proposed methodology follows a two-stage approach to efficiently segment kidney and tumour structures. The process begins with voxel sparsification, which reduces computational complexity by retaining only the most relevant voxels. Next, a region of interest (ROI) identification step is employed to localise potential areas containing kidney and tumour structures. Finally, a full segmentation stage refines the predictions, generating detailed and accurate segmentations of the target regions. An illustration of these steps can be found in Fig.~\ref{fig:method}.

\subsection*{Dataset}

For this study, we used publicly available images of renal cancer patients collected for the 2023 Kidney Tumour Segmentation challenge (KiTS23)~\cite{kits23}. This dataset includes patients who underwent cryoablation, partial nephrectomy or radical nephrectomy for suspected renal malignancy between 2010 and 2022 at an M Health Fairview medical center. Each case corresponds to the contrast-enhanced preoperative scan in either corticomedullary or nephrogenic phase. The challenge dataset was formed by a total of 599 CT scans, from which 489 were allocated to the training set and 110 to the test set. Our study had access only to the training subset, therefore only 489 CT scans were used for training and validation. The challenge annotations (ground truth segmentations) for kidney and masses (tumours and cysts) were used, in particular the version using a majority vote amongst several readers. Ethical approval for this analysis was not required, as stated in the license accompanying the open-access dataset.

In order to homogenise the voxel sizes of all the cases used prior to training, we performed a resampling to a fixed isotropic voxel size of the scans. Specifically, we used reference values used in recent editions of the KiTS challenge:
\begin{itemize}
\item Low resolution: (1.99, 1.99, 1.99)\,mm for the first stage (ROI finder, detection) as proposed in \cite{10.1007/978-3-031-54806-2_10}.
\item High resolution: (0.78, 0.78, 0.78)\,mm for the second stage (full segmentation) as proposed in \cite{uhm2023exploring3dunettraining, kits23winner, 10.1007/978-3-031-54806-2_3}.
\end{itemize}

\subsection*{Voxel Sparsification}

Sparsification requires voxels to be ignored from input images if they are deemed unimportant to the segmentation task. This step is crucial for reducing computational complexity, minimising memory usage, and enhancing model focus on relevant anatomical structures. By eliminating redundant information, such as large black regions surrounding the patient, the model can better allocate its capacity to learn meaningful features for segmentation. For example, the large black regions of the image surrounding the patient, as shown in the examples from Fig.~\ref{fig:method}, are superfluous for the identification of the kidneys and masses. Besides, sparsification is a prerequisite for using Sparse Submanifold Convolutional Networks (SSCNs). Unlike traditional dense convolutional networks, which process all voxels equally, SSCNs efficiently leverage sparse data structures that focus computations only on non-zero elements.

\begin{figure}[tbh]
  \includegraphics[width=0.48\textwidth, height=0.35\textwidth]{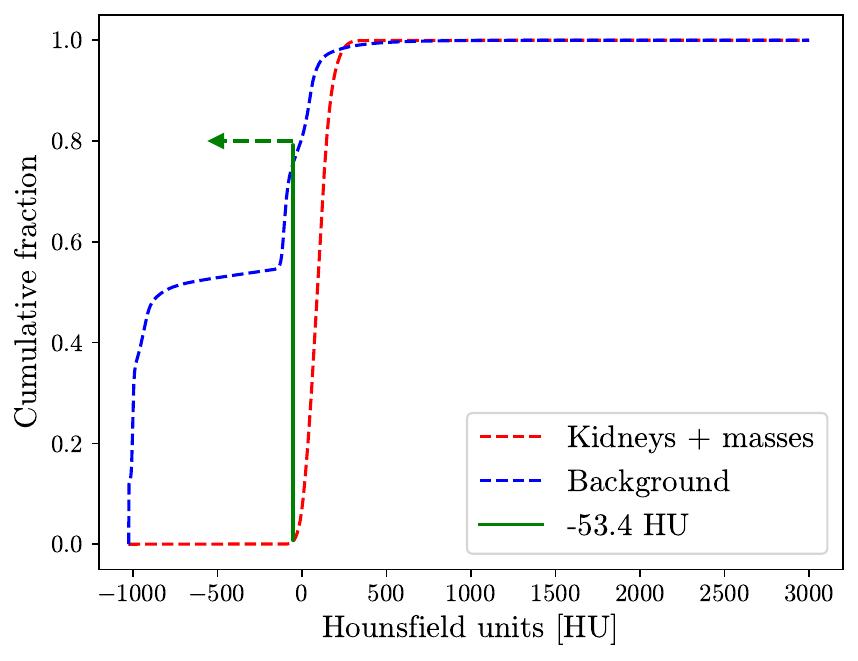}
  \includegraphics[width=0.48\textwidth, height=0.35\textwidth]{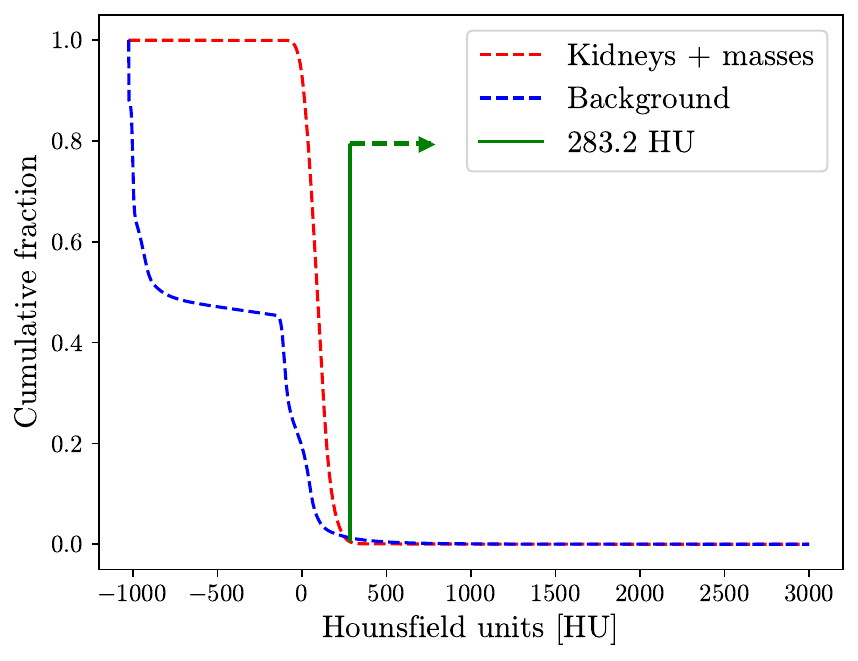}
  \caption{\textbf{Voxel sparsification based on Hounsfield Unit thresholds.} The plots show the cumulative fraction of voxels removed by applying a minimum (left) and maximum (right) threshold to the voxel intensity in Hounsfield Units (HU), for kidney and masses voxels (red) and all other voxels (blue); the green arrows indicate the threshold regions rejected in the sparsification process.}
  \label{fig:min_max_voxels}
\end{figure}

To determine the most effective sparsification strategy in the first stage (low-resolution ROI finder or detection stage), we optimised the range of Hounsfield Units (HU) within which voxels pertaining to kidneys and masses are retained. Instead of using predefined thresholds, we derived the sparsification range using percentiles of voxel intensities. Specifically, we computed the 0.5th and 99.5th percentiles of the HU values for all voxels inside the segmentations (kidneys and masses), ensuring that 99\% of the segmentation voxels are preserved. This approach yielded a threshold range of (-53.4, 283.2) HU. As shown in Fig.~\ref{fig:min_max_voxels}, this method retains approximately 99\% of segmentation voxels (red), while effectively removing 76.8\% of background voxels (blue), leading to a more efficient and targeted segmentation process.

For the second stage (high-resolution segmentation), we leverage the region of interest (ROI) obtained from the first stage rather than applying sparsification based on HU thresholds. 
This ensures that the segmentation network processes only the cropped region identified as relevant, reducing computational cost while maintaining fine-grained segmentation accuracy.

In our SSCN implementation, the Stage-1 HU filtering also determines the active voxel set for sparse convolution. Removing this filtering (or substantially widening the HU range) makes the Stage-1 input nearly dense at the full field of view, which exceeded memory limits in our experiments. We therefore represent non-sparse processing through the external dense baselines and the architecture-matched dense U-Net comparisons reported in the Results.

\subsection*{Data augmentations}
\label{sec:augmentations}

We used the Medical Open Network for AI (MONAI) framework~\cite{cardoso2022monaiopensourceframeworkdeep} to apply a series of data augmentations to increase the dataset diversity and model robustness. Affine transformations were used for spatial variability, including translations, rotations, and scaling, while random flipping along all three axes further diversified spatial orientation. We also applied intensity-based augmentations, such as random scaling and shifting of image intensities, to simulate varying imaging conditions. Gaussian noise and smoothing were added to mimic real-world acquisition noise and blurring. These augmentations were implemented within a unified MONAI pipeline, ensuring a structured and reproducible augmentation strategy for improved model training, and they follow usual practices widely used in the literature. Further details can be found in the Appendix. 

\subsection*{Network architecture}
\label{sec:networks}

\begin{figure*}[tbh]
   \centering
   \includegraphics[width=1.0\textwidth]{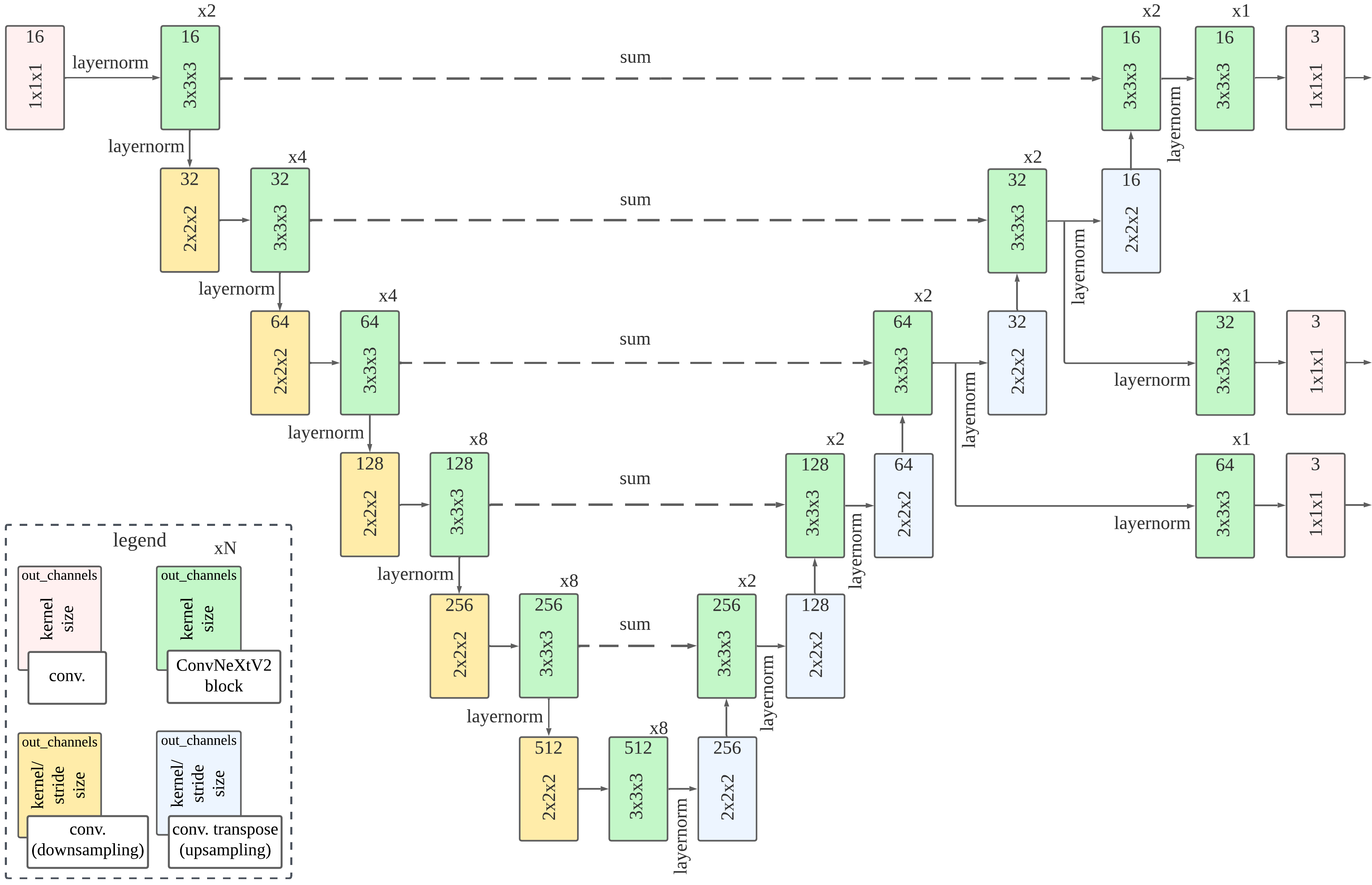}
   \caption{\textbf{The proposed 3D sparse U-Net architecture}. The network follows a hierarchical encoder-decoder structure with progressively increasing feature dimensions in the encoder and corresponding feature reduction in the decoder. Downsampling and upsampling operations are performed with convolution and transposed convolution layers, respectively, while skip connections are implemented via element-wise summation to maintain parameter efficiency.}
   \label{fig:unet}
\end{figure*}

We implemented a custom sparse voxel-based U-Net architecture for both Stage 1 (ROI detection) and Stage 2 (full-resolution segmentation), as illustrated in Fig.~\ref{fig:unet}. To efficiently process sparse volumetric data, we employed the MinkowskiEngine (v0.5.4)~\cite{choy20194d}, a specialised framework designed for sparse tensor operations. The MinkowskiEngine enables highly optimised submanifold sparse convolutions, significantly reducing computational overhead by focusing operations only on active (non-zero) voxel regions. However, the standard MinkowskiEngine does not support depth-wise convolutions, a critical component of our architecture. To address this limitation, we used a custom version{\footnote{\url{https://github.com/shwoo93/MinkowskiEngine}}}  of MinkowskiEngine, which incorporates a custom CUDA kernel for depth-wise convolutions, allowing our model to leverage this efficiency in both encoder and decoder layers. The model was implemented in Python version 3.10.5 using PyTorch version 2.0.0.

The architecture follows a fully convolutional encoder-decoder structure with ConvNeXtV2~\cite{woo2023convnextv2codesigningscaling} blocks as its core units. The network begins with a stem layer, projecting the input channels to the initial feature dimension, composed of a single \texttt{MinkowskiConvolution} (kernel size $1\times1\times1$) followed by \texttt{MinkowskiLayerNorm}. The encoder progressively downsamples the input while expanding feature dimensions. It consists of six hierarchical stages with increasing feature sizes: (16, 32, 64, 128, 256, 512). Each stage contains multiple \texttt{ConvNeXtV2} blocks with depths (2, 4, 4, 8, 8, 8). All convolutions in these blocks use a kernel size of $3\times3\times3$ and maintain spatial resolution. Downsampling occurs between stages using \texttt{MinkowskiConvolution} layers (kernel size $2\times2\times2$, stride $2\times2\times2$), halving spatial resolution while doubling feature dimensions. This hierarchical structure ensures a sufficiently large receptive field in every dimension. The decoder mirrors the encoder, progressively upsampling feature maps while reducing feature dimensionality: (512, 256, 128, 64, 32, 16). The decoder contains five stages, comprising two \texttt{ConvNeXtV2} blocks. Upsampling is performed using \texttt{MinkowskiConvolutionTranspose} layers (kernel size $2\times2\times2$, stride $2\times2\times2$), doubling spatial dimensions while halving feature channels. Skip connections are implemented through element-wise summation instead of concatenation, reducing memory footprint and parameter count. Each decoder stage has a classification sub-network consisting of a \texttt{MinkowskiLayerNorm}, a \texttt{ConvNeXtV2} block, and a final \texttt{MinkowskiConvolution} to produce predictions at multiple resolutions. Each classification sub-network has three output channels, corresponding to the three classes of the problem: all foreground combined (kidneys + masses), masses (tumour + cyst), and tumour only. The final activation function is sigmoid, enabling multi-label segmentation. The model has a total of 27,473,696 parameters.

Deep supervision is applied in the last three stages of the decoder, where the loss function follows the weighted formulation:
\begin{equation}
    \text{loss} = \sum_{i=0}^{3} \frac{1}{2^i} \text{Dice\_loss}(\text{pred}, \text{target}^{\downarrow})
\end{equation}
where the weight $\frac{1}{2^i}$ decreases for each sublevel $i$, corresponding to smaller spatial resolutions. This strategy, inspired by~\cite{kits23winner}, ensures gradient propagation at different scales. Target labels are downsized (if necessary) to match the corresponding output size through auxiliary label generation via a \texttt{MinkowskiAvgPooling} layer (indicated as $\text{target}^{\downarrow})$). The loss function used for training is the Dice loss~\cite{milletari2016vnet, 10.1007/978-3-319-67558-9_28}, which is defined as:

\begin{equation}
 \text{Dice\_loss}(\text{pred}, \text{target}) = 1 - 2\frac{pred\cap target}{pred + target + \epsilon}
\end{equation}

This loss formulation is particularly suitable for the present task because the target structures (especially tumours) cover a small fraction of the total volume, leading to strong class imbalance at the voxel level. By directly optimising overlap, Dice loss reduces the dominance of background voxels compared with voxel-wise classification losses, and improves the sensitivity for small lesions. In addition, the weighted deep-supervision scheme encourages stable gradient propagation across decoder resolutions, which was important for the convergence of both the ROI finder and the full-resolution segmentation networks.

Hyperparameter optimisation was guided using Optuna~\cite{optuna_2019}. The optimiser used is AdamW~\cite{loshchilov2019decoupled} with a learning rate of $5\times10^{-4}$, weight decay of $1\times10^{-5}$, and $(\beta_1, \beta_2) = (0.9, 0.95)$. Training is performed with a batch size of 1 and gradient accumulation over 8 batches. The total training spans 500 epochs per fold, beginning with a single epoch of warm-up, followed by 99 epochs at a fixed learning rate, and concluding with 400 epochs of cosine annealing scheduling, gradually reducing the learning rate to zero. Weight initialisation follows a truncated normal distribution for convolutional kernels, while biases in normalisation layers are initialised to zero, ensuring stable training.

\subsection*{Stage 1: ROI finder (detection)}
\label{sec:roi_finder}
The first stage of our segmentation pipeline focuses on detecting the region of interest (ROI). This step aims to localise kidney and mass structures while efficiently minimising computational overhead.

Our custom U-Net architecture was employed for this task, trained to classify each voxel as either signal (kidney/tumour/cyst) or background using a voxel-wise probability thresholding approach.

Due to the limited number of images available for training, we performed a 5-fold cross-validation procedure as follows: after shuffling, the dataset was divided into five equal parts, where four were used for training, and one for validation/testing; this procedure is repeated five times, each time employing a different part for validation/testing. In each step of the 5-fold cross-validation, a new network instance is trained independently on the four corresponding remaining parts.

In order to define the ROI region, voxels with a prediction probability above 0.1 for any class were retained, and the ROI was expanded using MONAI’s elliptical dilation filter (of size 11 voxels) to compensate for minor inaccuracies. Specifically, this step ensures that potentially relevant segmentation regions missed in the low-resolution step are captured before proceeding to full-resolution segmentation.

\subsection*{Stage 2: Full segmentation}
\label{sec:full_segmentation}

The second stage of our segmentation pipeline refines the results obtained from the ROI finder by performing full-resolution segmentation on cropped high-resolution images. This stage aims to achieve finer anatomical segmentation by leveraging the detailed voxel representation within each detected region of interest.
The high-resolution images were cropped using the ROI results from Stage 1. To ensure spatial consistency, small disconnected regions ($<$50 voxels) were removed from the predicted ROIs. Connected component analysis was then applied to identify separate anatomical structures, allowing each high-resolution image to be split into $K$ independent components. Each component was processed separately, reducing the memory footprint and enabling efficient training by working with smaller images that fit within GPU memory constraints.

Each identified component was then used to crop the corresponding high-resolution scan, resulting in only the relevant anatomical structures within those components being retained for further processing. This approach allowed for a more focused segmentation task, reducing the amount of irrelevant background information while preserving the fine details necessary for accurate segmentation. This refinement step ensured that more than 99\% of the original non-background voxels were retained within the final cropped high-resolution images.

The same custom U-Net architecture shown in Figure~\ref{fig:unet} was employed for this stage, trained following the same 5-fold cross-validation scheme. The final segmentation masks were reconstructed by mapping each high-resolution prediction back to the original scan space, ensuring spatial consistency with the input images. This two-stage approach, combining coarse ROI detection with high-resolution refinement, enabled efficient processing while preserving fine anatomical details, leading to improved segmentation performance.

\subsection*{Baselines}
\label{sec:baselines}

To benchmark the proposed sparse architecture against representative alternatives used for 3D medical image segmentation, we compared it with two external baselines: a task-specific self-configuring dense framework (nnU-Net) and a zoom-out/zoom-in foundation model (SegVol). Both were evaluated with the same grouped metrics used throughout this work: binary Dice similarity coefficients for kidneys + masses (foreground vs.\ background), tumour + cyst, and tumour alone.

\paragraph{nnU-Net.}
We trained nnU-Net~\cite{isensee2021nnunet} on the same KiTS23 dataset using identical 5-fold cross-validation splits. The nnU-Net is a self-configuring framework for deep learning-based biomedical image segmentation that automatically adapts its preprocessing pipeline, network architecture, and training schedule to a given dataset without manual intervention. We used nnU-Net v2 in its \texttt{3d\_fullres} configuration, which selected a PlainConvUNet architecture with 31.2\,M parameters, comparable to the 27.5\,M parameters of our sparse model. The framework automatically determined the training parameters, including a patch size of $128\times128\times128$ voxels, a batch size of 2, and a target spacing of $(1.0, 0.78, 0.78)$\,mm. Training was performed for 1000 epochs per fold with stochastic gradient descent and Nesterov momentum. Inference was performed with nnU-Net's standard sliding-window procedure without test-time augmentation.

\paragraph{SegVol.}
We also evaluated SegVol~\cite{du2024segvol}, a ViT\cite{dosovitskiy2021imageworth16x16words}-based foundation model (180.9\,M parameters) pre-trained on over 90,000 CT volumes spanning 200 anatomical categories, including kidney and renal lesion structures. SegVol was evaluated using the released pre-trained weights only, without task-specific fine-tuning. We used text-prompted inference with the prompts ``kidney'', ``kidney tumor'', and ``kidney cyst'' from the model's M3D-Seg vocabulary. Inference employed the model's zoom-out/zoom-in strategy, where an initial low-resolution prediction guides a spatially refined sliding-window pass with box prompts. The same five folds were used to group cases for evaluation, and the reported values correspond to the fold-wise averages computed with the same metrics as for the trained models.

\subsection*{Training and validation losses}
\label{sec:losses}

\begin{figure*}[htb]
  \centering
  \includegraphics[width=1.0\textwidth,trim=0 0 0 0,clip]{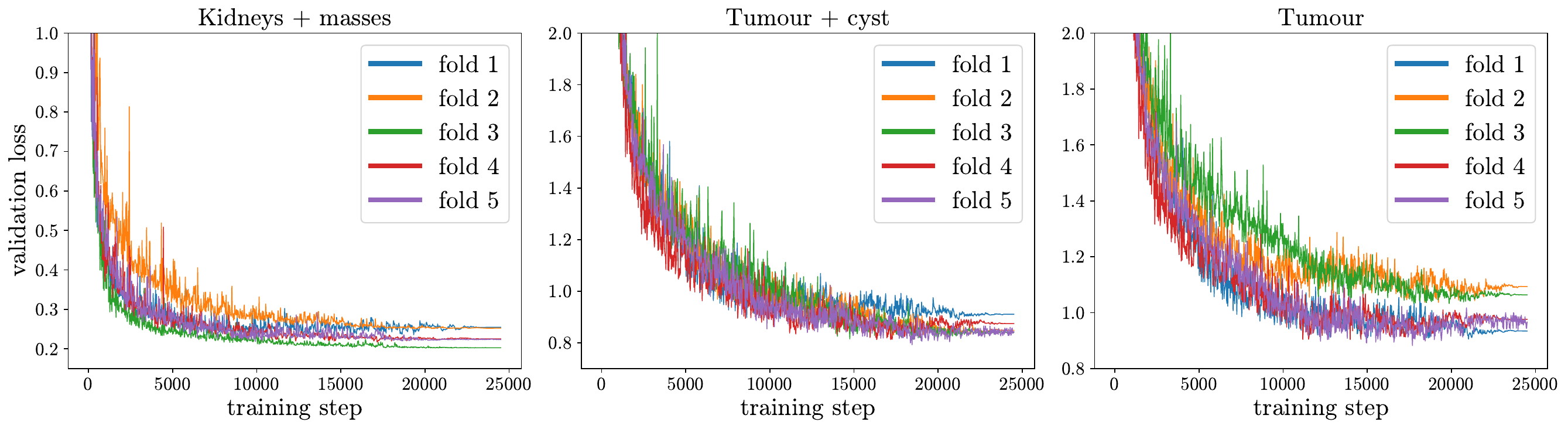} 
  \vskip 0.5cm
  \includegraphics[width=1.0\textwidth,trim=0 0 0 0,clip]{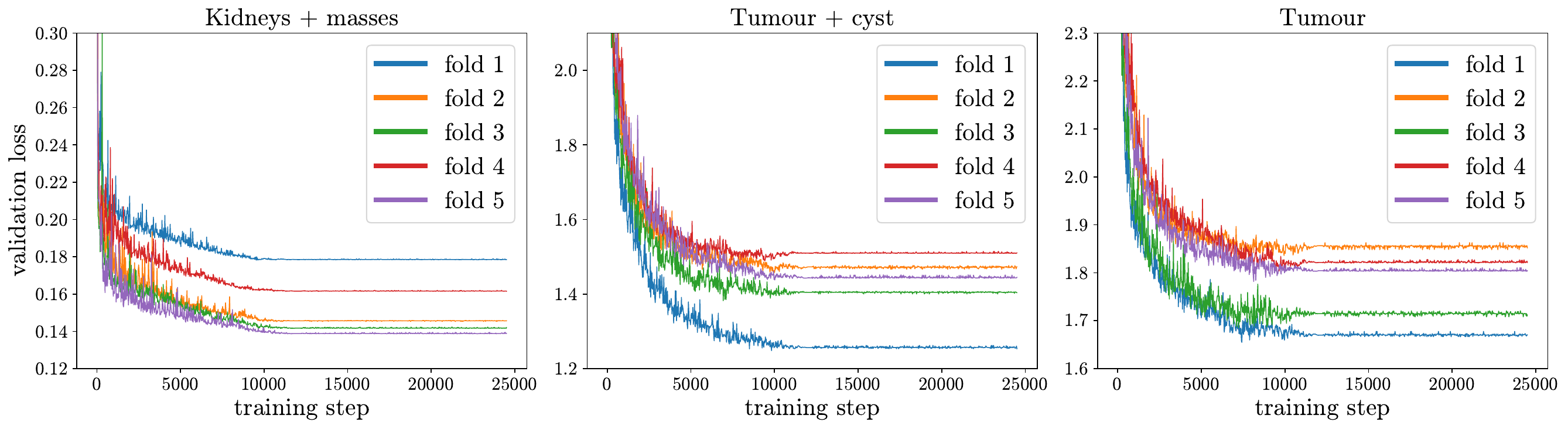} 
  \caption{\textbf{Validation losses for the two-stage sparse segmentation framework}. The top row corresponds to Stage 1 and the bottom row to Stage 2; each column shows the accumulated Dice loss across all deep supervision steps for a different output: kidneys + masses (left), tumour + cyst (middle), and tumour only (right). Different colours indicate different folds.}
  \label{fig:losses}
\end{figure*}

The validation losses for both stages are illustrated in Figure~\ref{fig:losses}. The plots show smooth convergence in all cases, with the high-resolution setup converging more quickly. One possible explanation is that, although high-resolution images are expected to provide more detail than low-resolution ones and are therefore more complex
, they are segmented into distinct (and likely simpler) connected components that are processed independently. This separation may also contribute to the higher Dice loss values observed in the high-resolution case for the masses outputs (Tumour + cyst and Tumour only), as some components might lack labels for one or both of the masses outputs. In such cases, the Dice similarity coefficient is 0, resulting in a Dice loss of 1 (the maximum possible value, given the small smoothing constant in the nominator and denominator). Overall, the smooth convergence observed across folds (Fig.~\ref{fig:losses}) is consistent with the use of Dice loss with weighted deep supervision, which is well matched to the class imbalance and multi-scale nature of this segmentation problem.

\section*{Results}

\begin{figure*}[thb!]
  \centering
  \includegraphics[width=1.0\textwidth,trim=0 120 0 0,clip]{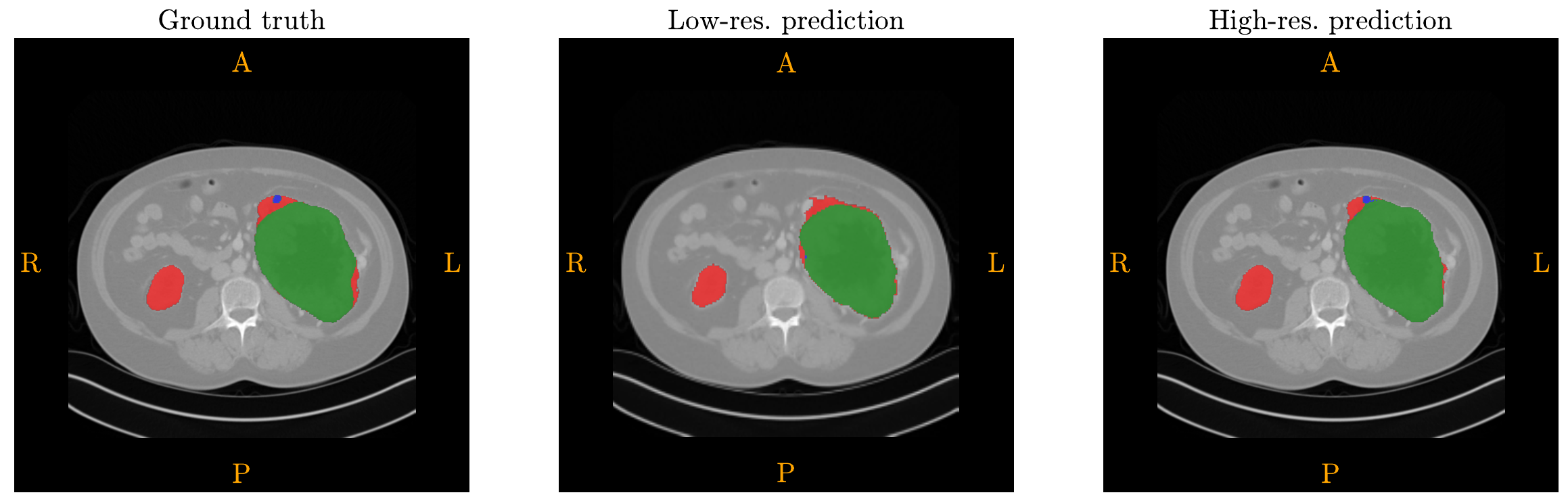} 
  \vskip 0.5cm
  \includegraphics[width=1.0\textwidth,trim=0 120 0 0,clip]{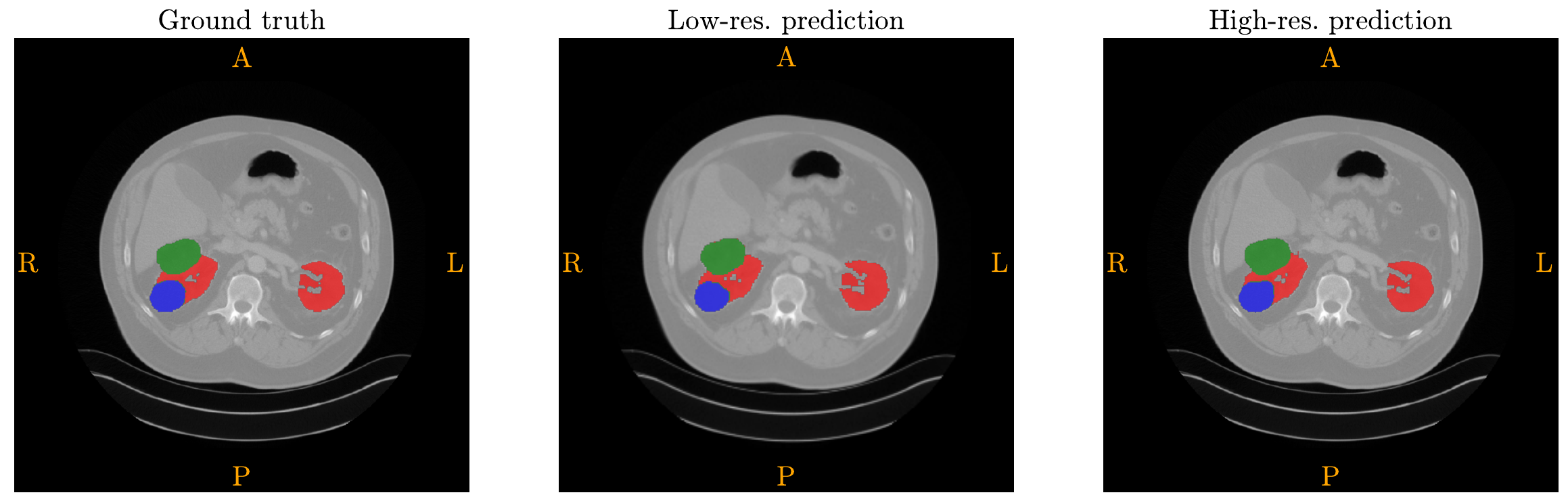} 
  \vskip 0.5cm 
  \includegraphics[width=1.0\textwidth]{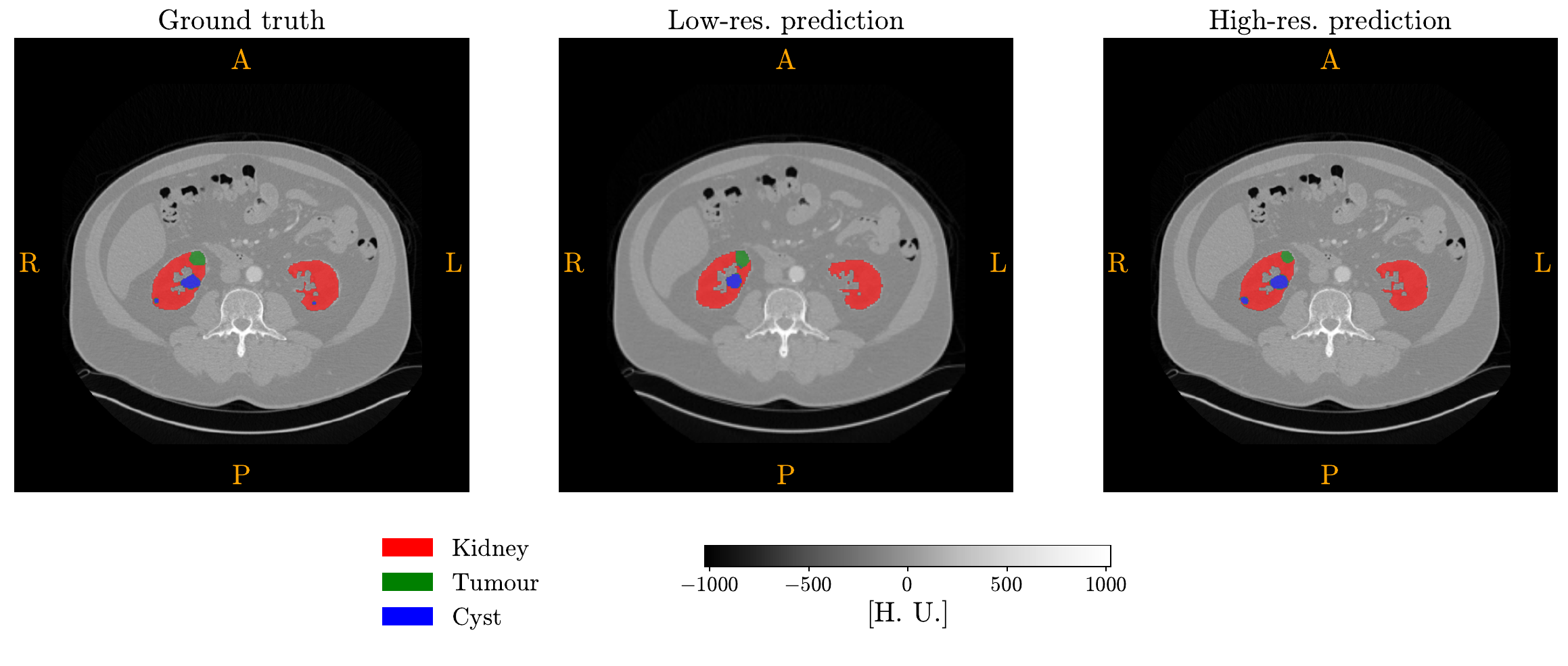}
  \caption{\textbf{Qualitative examples of low- and high-resolution segmentation outputs}. From left to right, the columns show: (first) a 2D slice from the original high-resolution scan with ground-truth segmentations, (second) the predicted segmentation from Stage 1 on the low-resolution scan, and (third) the predicted segmentation from Stage 2 on the high-resolution scan. Each row corresponds to a different case (patient).}
  \label{fig:results_roi}
\end{figure*}

This section presents the segmentation results for both stages of our method: low-resolution and high-resolution segmentations. Performance is evaluated using the Dice similarity coefficient (DSC), computed through 5-fold cross-validation at each stage. Fig.~\ref{fig:results_roi} shows visual examples from individual patients, highlighting the improvement from low- to high-resolution segmentation. Additionally, Tab.~\ref{tab:performance} reports the mean class-wise performance across folds. These examples demonstrate how the coarse Stage 1 output enables Stage 2 to recover fine details with high fidelity, including small cystic and tumoural structures. More details of the results for the low and high-resolution networks are given below.

\begin{table*}[htb]
    \centering
    \caption{\textbf{Segmentation performance of all evaluated models across folds in terms of Dice similarity coefficient}. To ensure a fair comparison, all metrics are computed in the original KiTS23 image space (native voxel grid / voxel spacing of each case), after mapping each model prediction back from its internal processing resolution. Row ``All'' corresponds to the arithmetic mean (macro-average) of the three reported outputs: kidneys + masses, tumour + cyst, and tumour. Bold values indicate the average performance across folds. *SegVol was evaluated using the
released pre-trained weights only (zero-shot mode), without task-specific fine-tuning. }
    \label{tab:performance}
    \renewcommand{\arraystretch}{1.2}
    \begin{adjustbox}{max width=1.1\textwidth,center}
    \begin{tabular}{llcccccc}
        \toprule
        \multirow{2}{*}{\textbf{Network}} & \multirow{2}{*}{\textbf{Output}} & \multicolumn{5}{c}{\textbf{Folds}} & \multirow{2}{*}{\textbf{Average}} \\
        \cmidrule(lr){3-7}
        & & \textbf{Fold 1} & \textbf{Fold 2} & \textbf{Fold 3} & \textbf{Fold 4} & \textbf{Fold 5} & \\
        \midrule
        \multirow{4}{*}{nnU-Net~\cite{isensee2021nnunet}}
        & Kidneys + masses & 0.9666 & 0.9576 & 0.9702 & 0.9658 & 0.9741 & \textbf{0.9669} \\
        & Tumour + cyst & 0.8182 & 0.8274 & 0.8366 & 0.8429 & 0.8580 & \textbf{0.8366} \\
        & Tumour & 0.7793 & 0.7544 & 0.7827 & 0.8200 & 0.8461 & \textbf{0.7965} \\
        & All (macro-average) & 0.8547 & 0.8464 & 0.8632 & 0.8762 & 0.8927 & \textbf{0.8666} \\
        \midrule
        \multirow{4}{*}{SegVol~\cite{du2024segvol}}
        & Kidneys + masses & 0.9115 & 0.9012 & 0.9165 & 0.9152 & 0.9256 & \textbf{0.9140*} \\
        & Tumour + cyst & 0.2182 & 0.2181 & 0.1787 & 0.2300 & 0.2472 & \textbf{0.2184*} \\
        & Tumour & 0.2105 & 0.1998 & 0.1381 & 0.2031 & 0.2247 & \textbf{0.1953*} \\
        & All (macro-average) & 0.4467 & 0.4397 & 0.4111 & 0.4495 & 0.4658 & \textbf{0.4426*} \\
        \midrule
        \multirow{4}{*}{Low-resolution (ours)}
        & Kidneys + masses & 0.9409 & 0.9416 & 0.9589 & 0.9496 & 0.9524 & \textbf{0.9487} \\
        & Tumour + cyst & 0.7353 & 0.7577 & 0.7607 & 0.7445 & 0.7556 & \textbf{0.7508} \\
        & Tumour & 0.7283 & 0.6672 & 0.6788 & 0.7092 & 0.7034 & \textbf{0.6974} \\
        & All (macro-average) & 0.8015 & 0.7888 & 0.7995 & 0.8011 & 0.8038 & \textbf{0.7989} \\
        \midrule
        \multirow{4}{*}{High-resolution (ours)}
        & Kidneys + masses & 0.9526 & 0.9514 & 0.9676 & 0.9580 & 0.9610 & \textbf{0.9581} \\
        & Tumour + cyst & 0.8313 & 0.8576 & 0.8671 & 0.8626 & 0.8657 & \textbf{0.8568} \\
        & Tumour & 0.8067 & 0.7930 & 0.7734 & 0.8143 & 0.8292 & \textbf{0.8033} \\
        & All (macro-average) & 0.8635 & 0.8673 & 0.8694 & 0.8783 & 0.8853 & \textbf{0.8728} \\
        \bottomrule
    \end{tabular}
    \end{adjustbox}
\end{table*}

\subsection*{Low-resolution network}

As described above, the purpose of the low-resolution network is to find the ROIs containing the kidneys, tumours and cysts, that form the input to the high-resolution network. It achieves a high DSC of $94.9\%$ for the kidney + masses output, confirming its effectiveness in reliably identifying ROIs. The detection performance for tumours + cysts and tumours alone reaches $75.1\%$ and $69.7\%$, respectively. While these predictions are not expected to be used directly for final segmentation, they are instrumental in focusing the second stage on relevant anatomical regions.

\subsection*{High-resolution network and baseline comparison}

Table~\ref{tab:performance} reports the results of both external baselines described in the Methods section, evaluated on the same cross-validation splits and metrics as our model. The nnU-Net baseline achieves an average DSC of $96.7\%$ for kidneys + masses, $83.7\%$ for tumour + cyst, and $79.7\%$ for tumour alone. Our high-resolution sparse model obtains $95.8\%$, $85.7\%$, and $80.3\%$ for the same categories, respectively. Therefore, relative to nnU-Net, our method is slightly lower for kidneys + masses but numerically higher for the lesion-focused outputs (tumour + cyst and tumour alone), with a comparable overall macro-average (``All'': $87.3\%$ vs.\ $86.7\%$).

SegVol, evaluated in zero-shot mode using pre-trained weights only, achieves a kidney + masses DSC of $91.4\%$, confirming its ability to localise large organs without task-specific training. However, its performance on finer-grained targets is substantially lower ($21.8\%$ for tumour + cyst and $19.5\%$ for tumour alone), yielding an overall macro-average of $44.3\%$. This gap relative to both our method and nnU-Net is consistent with the difficulty of zero-shot segmentation of small, heterogeneous renal lesions. While beyond the scope of this article, the performance for tumour and cysts would possibly improve with task-specific fine-tuning.

These results show that the proposed sparse methodology achieves tumour segmentation accuracy comparable to a state-of-the-art patch-based dense frameworks, while the computational comparison in the following section highlights its advantages regarding computational efficiency.

\subsection*{Error analysis by lesion burden}

\begin{figure*}[htb]
  \centering
  \includegraphics[height=0.2\textwidth]{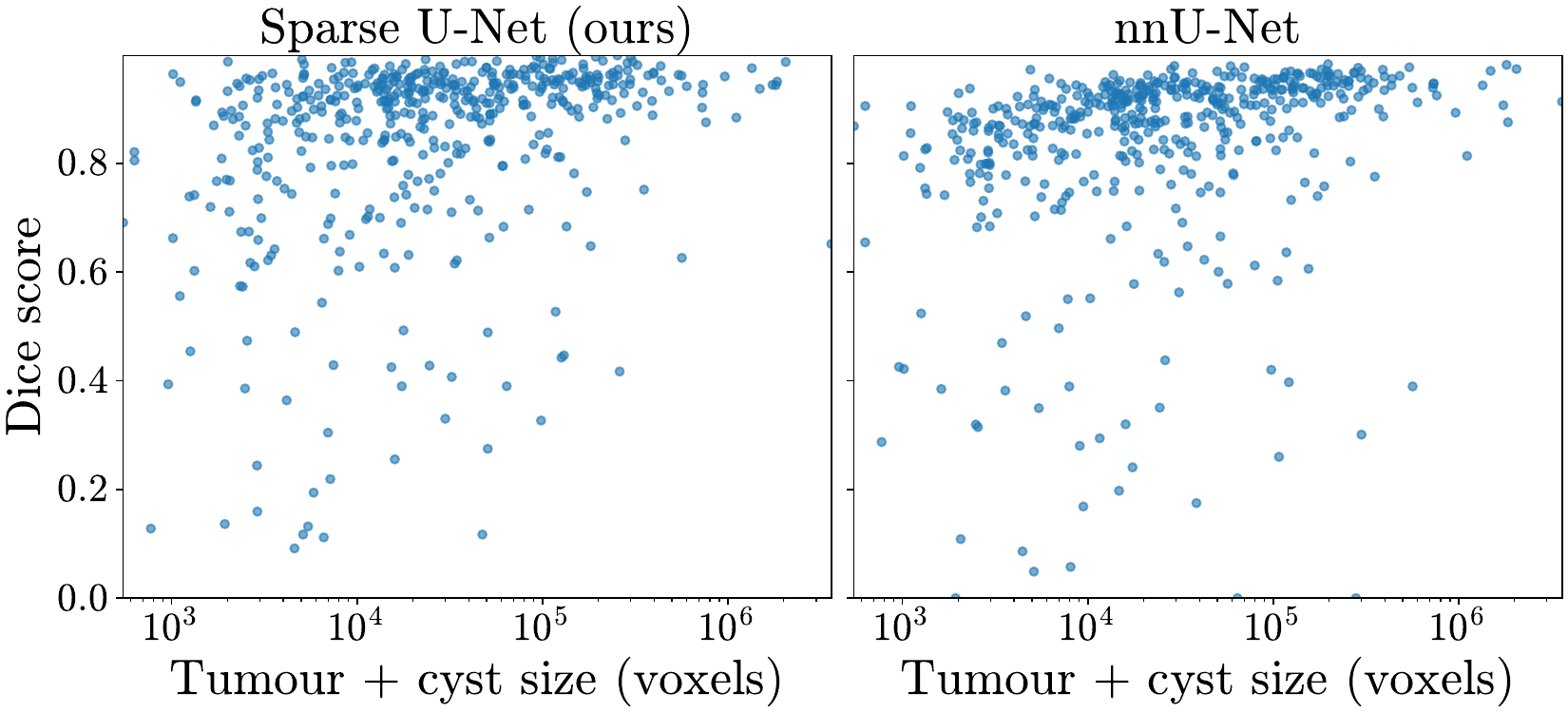}\hspace{0.03\textwidth}%
  \includegraphics[height=0.2\textwidth,]{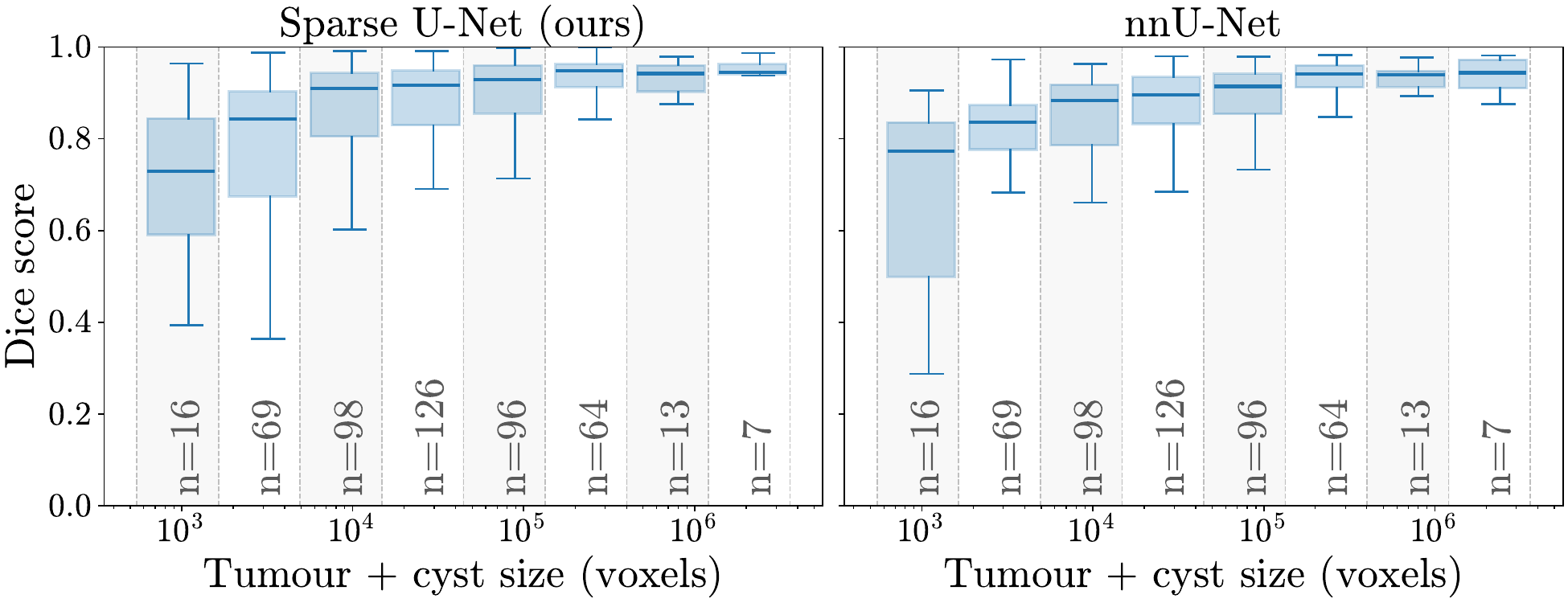}
  \caption{\textbf{Case-level error analysis as a function of lesion burden (tumour + cyst)}. Left: Scatter plots of tumour + cyst Dice score versus ground-truth tumour + cyst size (voxels, log scale) for the proposed sparse U-Net (left) and nnU-Net (right). Right: The same data summarised in logarithmic size bins using boxplots, with the number of cases per bin ($n$) indicated on the x-axis. Cases from all 5 validation folds are pooled. Tumour + cyst size is measured per case from the ground-truth annotations in the original KiTS23 image space (i.e. the common evaluation space used for both methods).}
\label{fig:error_analysis_size}
\end{figure*}

To provide a brief error analysis for challenging cases, we examined tumour + cyst segmentation performance as a function of case-level lesion burden (ground-truth tumour + cyst volume in voxels), pooling all validation cases across the 5 folds for our sparse model and nnU-Net. SegVol was not included in this analysis because its tumour + cyst Dice score is substantially lower overall (Table~\ref{tab:performance}), which would make the comparison less informative for the two methods that are competitive on this output.
Figure~\ref{fig:error_analysis_size} shows the case-level scatter-plot distribution and summarises the same relationship with boxplots and per-bin case counts. Both methods show lower and more variable tumour + cyst Dice scores for cases with smaller tumour + cyst burden, whereas performance becomes more stable as lesion volume increases. Across the size range, the proposed sparse model remains broadly comparable to nnU-Net, consistent with the fold-averaged results reported in Tab.~\ref{tab:performance}.

\subsection*{Computational Performance}

Details of the computational performance of the low-resolution and high-resolution networks are given in Tab.~\ref{tab:computational_performance} when using a CPU compared to a set of different GPUs. The Sparse U-Net results use depth-wise convolutions on GPU, leveraging a custom CUDA kernel from an extended MinkowskiEngine implementation (\url{https://github.com/shwoo93/MinkowskiEngine}); however, this implementation lacks CPU support, so for CPU inference, all depth-wise convolutions were converted into MinkowskiEngine's built-in channel-wise convolutions, yielding the same predictions but being less efficient on GPU. For the high-resolution case, since a scan may be divided into $K$ disconnected components that are processed independently
, the reported results first sum the metric (inference time or VRAM) across all components of each scan, then take the average across all scans.

As shown in Tab.~\ref{tab:computational_performance}, dense models frequently ran out of memory, especially at higher resolutions and batch sizes. In configurations where both models ran successfully, the sparse U-Net achieved reductions in inference time ranging from approximately 33\% to 60\%, and VRAM usage was consistently lower, typically by a factor of 2 to 4. These results were consistent across both low- and high-resolution settings and across various GPUs, including V100 and A100 architectures. The much smaller memory usage makes it possible to perform the segmentation in 3D at full resolution. 
Table~\ref{tab:computational_performance} also includes the two external baselines. Across the full set of GPU configurations, nnU-Net is the most computationally demanding baseline: it runs out of memory on the V100 16\,GB card and, on both A100 GPUs, requires substantially more inference time and VRAM than our sparse model due to dense overlapping sliding-window inference with $128^3$ patches. In contrast, SegVol runs on all tested GPUs and uses the least VRAM (approximately 2\,GB, nearly independent of GPU type) thanks to its fixed $32{\times}256{\times}256$ sliding-window design, but it remains slower than our sparse model and, as shown in the previous section, it is markedly less accurate for lesion segmentation. The trends for our architecture-matched sparse and dense U-Nets are also consistent across CPU and GPU settings: the sparse implementation is systematically faster and more memory-efficient, while the dense implementation exhibits frequent OOM failures, especially on the V100 and in batch-size-2 configurations.

\begin{table*}[htb]
    \centering
    \caption{\textbf{Comparison of average per-volume inference time and peak VRAM for all evaluated models}. The Sparse and Dense U-Net models are tested at both low and high resolutions with batch sizes of 1 and 2. nnU-Net uses sliding-window inference with $128^3$ patches at its auto-configured target spacing without test-time augmentation. SegVol uses text-prompted zoom-out/zoom-in inference with $32{\times}256{\times}256$ sliding windows and pre-trained weights only (zero-shot evaluation). Both baselines are restricted to batch size~1 by their inference design, which processes a single volume at a time. The ``Dense U-Net'' model is the equivalent dense implementation of the sparse model. 
    The CPU used for testing is an Intel(R) Xeon(R) CPU E5-2698 v4 @ 2.20GHz. Note that the large variability in size (number  of 2D slices) between cases results in large errors. OOM: out of memory.}
    \label{tab:computational_performance}
    \begin{adjustbox}{max width=\textwidth,center}
    \renewcommand{\arraystretch}{2} 
    \begin{tabular}{llc ccccccc}
        \toprule
        \multirow{2}{*}{\textbf{Model}} & \multirow{2}{*}{\textbf{Inference space}} & \multirow{2}{*}{\textbf{Batch size}} & \multicolumn{4}{c}{\textbf{Average inference time (s)}} & \multicolumn{3}{c}{\textbf{Average peak VRAM (GB)}} \\
        \cmidrule(lr){4-7} \cmidrule(lr){8-10}
        & & & \textbf{CPU} & \textbf{V100 16GB} & \textbf{A100 40GB} & \textbf{A100 80GB} & \textbf{V100 16GB} & \textbf{A100 40GB} & \textbf{A100 80GB} \\
        \midrule
         nnU-Net~\cite{isensee2021nnunet} &  Auto-configured resampled &  1 &  $592.48 \pm 296.50$ &  OOM &  $7.88 \pm 4.34$ &  $7.85 \pm 4.33$ &  OOM &  $5.90 \pm 1.60$ &  $5.88 \pm 1.59$\\
        \midrule
         SegVol~\cite{du2024segvol} &  Native image space &  1 & $305.86 \pm 320.57$ &  $1.24 \pm 1.03$ &  $1.22 \pm 1.03$ &  $1.24 \pm 1.04$ &  $1.99 \pm 0.30$ & $1.96 \pm 0.29$ &  $1.96 \pm 0.29$ \\
        \midrule
        \multirow{4}{*}{Sparse U-Net (ours)} & \multirow{2}{*}{Low-resolution}  & 1 & 29.38 $\pm$ 15.52  & 0.39 $\pm$ 0.14 & 0.23 $\pm$ 0.08 & 0.22 $\pm$ 0.07 & 2.87 $\pm$ 1.74  & 2.85 $\pm$ 1.76 & 2.84 $\pm$ 1.76 \\
        &  & 2 & 34.41 $\pm$ 11.90 & OOM & 0.21 $\pm$ 0.06 & 0.20 $\pm$ 0.06 & OOM & 5.58 $\pm$ 2.73 & 5.60 $\pm$ 2.74 \\
        & \multirow{2}{*}{High-resolution} & 1 & 32.66 $\pm$ 18.89 & 0.47 $\pm$ 0.15 & 0.29 $\pm$ 0.13 & 0.29 $\pm$ 0.13 & 3.25 $\pm$ 1.78 & 3.32 $\pm$ 1.76 & 3.29 $\pm$ 1.75 \\
        &  & 2 & 53.68 $\pm$ 25.85 & OOM & 0.25 $\pm$ 0.08 & 0.25 $\pm$ 0.07 & OOM & 6.69 $\pm$ 3.42 & 6.70 $\pm$ 3.45 \\
        \midrule
        \multirow{4}{*}{Dense U-Net (ours)} & \multirow{2}{*}{Low-resolution}  & 1 & 47.55 $\pm$ 25.21  & OOM & 0.37 $\pm$ 0.19 & 0.37 $\pm$ 0.19 & OOM & 8.70 $\pm$ 5.08 & 8.70 $\pm$ 5.09 \\
        &  & 2 & 66.85 $\pm$ 25.28 & OOM & OOM & 0.50 $\pm$ 0.21 & OOM & OOM & 23.62 $\pm$ 9.79 \\
        & \multirow{2}{*}{High-resolution} & 1 & 51.42 $\pm$ 28.45 & OOM & 0.43 $\pm$ 0.20 & 0.44 $\pm$ 0.20 & OOM & 10.28 $\pm$ 5.40  & 10.28 $\pm$ 5.39 \\
        &  & 2 & 63.26 $\pm$ 29.38 & OOM & OOM & 0.53 $\pm$ 0.21 & OOM & OOM & 25.45 $\pm$ 11.30 \\
        \bottomrule
    \end{tabular}
    \end{adjustbox}
\end{table*}


\section*{Discussion}

In this paper, we demonstrated the feasibility and potential for using submanifold sparse convolutional networks (SSCN) for the segmentation of organs and tumours in CT images, in particular we applied it to kidneys, kidney tumours and cysts as a proof of concept use case. To the best of our knowledge, this is the first time that such SSCNs have been applied to the segmentation of CT images, and only one more example exists in the literature in which the Minkowski Engine has been applied to medical imaging at all~\cite{Li2022}.

We proposed a new methodology that deploys such SSCNs in two stages, starting with a process of voxel sparsification applied to the original images. Then a SSCN was trained on a low-resolution version of the images to perform an initial ROI finding or detection, which obtains a good accuracy of
$\sim$\,95\% to select kidneys and masses. It is worth noting that this first stage can be used as a stand-alone and fast detection method by itself.  

Afterwards, in a second stage, the ROIs found are processed at full resolution to classify the selected voxels into the three categories of interest, with Dice similarity coefficients of $\sim$\,96\% for kidneys and masses, $\sim$\,86\% for tumour + cyst, and $\sim$\,80\% for tumours alone. The results achieved are competitive with other more established methods (typically based on dense U-Nets)~\cite{heller2023kits21challengeautomaticsegmentation, kits23winner, uhm2023exploring3dunettraining}, while significantly reducing the computational resources needed for training. In particular, compared to the results of the winners of the KiTS23 challenge \cite{kits23winner, kits23second}, our combined Dice similarity coefficient, calculated as average of the three classes (``All''), yields 87.3\%, well in agreement with the performance of the top teams in the challenge, reported on the training dataset to be 87\,-\,89\%. Particularly interesting is the comparison with the results of the method ranked second in the challenge \cite{kits23second}, as they also compared low and full-resolution training, reporting ranges of DSC of 86.5\,-\,88.7\% for the average of all classes, 97.3\,-\,97.9\% for kidney+masses, 84.0\,-\,85.8\% for tumour+cyst and 77.1\,-\,82.6\% for tumour only, and our results (87.3\%, 95.8\%, 85.7\% and 80.3\% respectively from the average of the cross-validation folds) lie comfortably within those ranges and comparable in the case of kidney + masses. Similarly, recent published work \cite{Hsiao2022Cmpb} has demonstrated competitive kidney–tumour segmentation using a dense U-Net variant, highlighting how our sparse approach offers a resource-efficient alternative. It is worth noting, for the fairness of the comparison, that we have not included any post-processing in our method yet, and that our results are strictly coming from the last training epoch, not the epoch with best validation accuracy.

In addition to the indirect comparisons with challenge results discussed above, we performed direct controlled comparisons with two external baselines (see Methods). The nnU-Net comparison shows that our sparse model achieves a comparable combined average DSC (87.3\% vs.\ 86.7\%), with a small decrease for kidneys + masses but numerically higher scores for the lesion-focused outputs (tumour + cyst and tumour alone). This result is particularly encouraging given that nnU-Net is widely regarded as a strong reference method in medical image segmentation, and that its configuration was auto-optimised for the KiTS23 dataset.

The SegVol comparison provides a complementary reference based on a zoom-out/zoom-in foundation-model inference strategy using pre-trained weights only. Despite achieving reasonable kidney + masses localisation (91.4\%), SegVol drops sharply on the finer-grained targets (21.8\% for tumour + cyst and 19.5\% for tumour alone), yielding a combined average of 44.3\%. This gap relative to both our method and nnU-Net highlights the difficulty of zero-shot segmentation for small, heterogeneous renal lesions, and underscores the value of task-specific training for this application.

Crucially, compared to equivalent dense U-Net architectures, our sparse network reduces inference time by up to 60\%, and memory consumption by as much as 75\% in some configurations. This not only enables processing of full-resolution volumes on limited hardware (e.g., GPUs with 16\,GB of memory), allowing deployment in clinical environments with limited computational resources, but also contributes to lower energy consumption during both training and deployment. As such, our method is better aligned with sustainable AI practices, helping to reduce the carbon footprint of large-scale medical imaging workflows~\cite{Climate}.

The computational comparison with the two baselines further highlights the efficiency of the sparse approach across the full range of tested hardware. On both A100 GPUs, our high-resolution sparse model is substantially faster and uses less VRAM than nnU-Net, while nnU-Net also fails on the V100 16\,GB configuration because of the memory requirements of dense overlapping patch-based inference. SegVol is comparatively memory-efficient and runs on all tested GPUs, but it remains slower than our sparse model and, as discussed above, achieves substantially lower lesion segmentation accuracy in zero-shot mode. In addition, the architecture-matched sparse-versus-dense comparison shows the same pattern on both CPU and GPU measurements: the sparse implementation is consistently faster and more memory-efficient, with the dense version frequently running out of memory in the more demanding settings.

The main limitations of this study are twofold. Firstly, the small size of the dataset, consisting of only 489 patients, can have a significant impact on the performance of the network, partly compensated by data augmentation. Secondly, as we performed this analysis outside the KiTS23 challenge, we did not have access to the labels corresponding to the test dataset. This means that we cannot directly compare our results to the ones reported in the leaderboard of this challenge~\cite{kits23}. This prevents us from investigating the generalisability of our model in unseen data, which could impact the performance of the model (results reported by the top-ranking teams of the challenge present a~5\% drop in performance in the test set). However, given the fact that we have not cherry-picked a best-performing epoch and that we used an average of 5-fold cross-validation, we are confident that the results we present here should generalise well in the test set. While Dice loss was selected for its suitability to highly imbalanced volumetric segmentation, a systematic comparison against alternative loss functions was outside the scope of the present study and is left for future work.

A controlled comparison against a single-stage high-resolution sparse model was not feasible in our setup, since the explicit ROI stage enables component-wise cropping and keeps full-resolution sparse inference within memory limits; evaluating alternative single-stage sparse designs under matched computational budgets is an important direction for future work.
Further work should include studies with larger datasets, ideally consisting of images from multiple centres, and exploring other tumours and organs as well in the future. We based this study on CT imaging as the intensity of the voxels is easily interpretable as measurements of different radiodensity, but similar sparsification processes could be explored for other image modalities such as MRI or PET. Adding a post-processing step, which seems to have been aided the performance of some of the top-ranking teams in the challenge, should be explored as well. 

In summary, this work demonstrates that sparse networks for the segmentation of organs and tumours in CT images are not only feasible but also achieve competitive accuracy while substantially reducing computational resources required.


\bibliography{biblio}

\section*{Acknowledgements}

The authors would like to thank Dr Thomas Buddenkotte for supporting our understanding of nnUNet-based automated segmentation, and the organisers of the KiTS23 challenge, especially Dr Nicholas Heller for his help using the public dataset.

\section*{Funding declaration}

No funding was required for this study.

\section*{Author contributions}

S.A.-M., L.H.W., A.A., and L.E.S contributed to the conception of the experiments and analysis. S.A.-M. and L.E.S gathered and prepared the dataset. S.A.-M. performed the training of the networks. S.A.-M., L.H.W., A.A., and L.E.S contributed to the discussion of the results and to the writing and reviewing of the manuscript.

\section*{Data availability}

The dataset used is accessible through the KiTS23 challenge website and linked GitHub repository \url{https://github.com/neheller/kits23}. 

\section*{ Code availability}

All code developed for this study is publicly available at \url{https://github.com/saulam/ai_cancer_research}.

\section*{Additional information}

The authors have no relevant financial or non-financial interests to disclose. No grants or financial support was relevant to this work. 

To facilitate the reproducibility of this study, all implementation details relevant to the data augmentations applied are listed here:

\begin{itemize}
    \item \textbf{Affine transformations}: 
    \begin{itemize}
        \item Random rotations:
        \begin{itemize}
            \item Around the x and y axes: Up to $\pi/36$ radians.
            \item Around the z-axis: Up to $\pi/8$ radians.
        \end{itemize}
        \item Random translations: Up to 30 voxels in-plane (x, y) and 5 voxels in the z direction.
        \item Random scaling: Up to 15\%.
    \end{itemize}
    \item \textbf{Flipping}: Applied randomly along the x, y, and z axes, each with a probability of 0.3.
    \item \textbf{Intensity variations}: 
    \begin{itemize}
        \item Random intensity scaling: factor 0.1, probability 0.3.
        \item Random intensity shifting: offset 5, probability 0.3.
    \end{itemize}
    \item \textbf{Noise and smoothing}: 
    \begin{itemize}
        \item Gaussian noise: Mean 0, standard deviation 1, probability 0.3.
        \item Gaussian smoothing (blurring):
        \begin{itemize}
            \item Random sigma values for x, y, and z axes: (0.25, 1.5).
            \item Probability of application: 0.3.
        \end{itemize}
    \end{itemize}
\end{itemize}

\end{document}